\documentclass[10pt,twocolumn,letterpaper]{article}

\usepackage{iccv}
\usepackage{times}
\usepackage{epsfig}
\usepackage{graphicx}
\usepackage{amsmath}
\usepackage{amssymb}
\usepackage{pifont}
\usepackage{enumitem}

\usepackage{color, colortbl}
\definecolor{LightCyan}{rgb}{0.88,1,1}

\usepackage{url}
\usepackage{graphicx}
\usepackage{amsmath}
\usepackage{booktabs}
\usepackage{multirow}
\usepackage{balance}
\usepackage{wrapfig,lipsum}

\usepackage{algorithm}
\usepackage{algpseudocode}
\usepackage{mathtools}
\usepackage{amsthm,amsmath,amssymb}
\usepackage{mathrsfs}
\usepackage{xcolor}
\usepackage[accsupp]{axessibility}

\newcommand{\eat}[1]{}

\usepackage{subcaption}

\usepackage[pagebackref=true,breaklinks=true,letterpaper=true,colorlinks,bookmarks=false]{hyperref}

\iccvfinalcopy 

\def\iccvPaperID{2513} 

\ificcvfinal\fi

\begin{document}

\title{Revisiting Domain-Adaptive 3D Object Detection by Reliable, Diverse and Class-balanced Pseudo-Labeling}


\author{Zhuoxiao Chen$^{1}$ \; Yadan Luo$^{1}$  \; Zheng Wang$^{2}$ \; Mahsa Baktashmotlagh$^{1}$ \;  Zi Huang$^{1}$ \\
$^1$The University of Queensland\quad $^2$University of Electronic Science and Technology of China \\
{\tt\small \{zhuoxiao.chen, y.luo, m.baktashmotlagh, helen.huang\}@uq.edu.au, zh\_wang@hotmail.com } \\
}

\maketitle
\ificcvfinal\thispagestyle{empty}\fi

\begin{abstract}
\noindent Unsupervised domain adaptation (DA) with the aid of pseudo labeling techniques has emerged as a crucial approach for domain-adaptive 3D object detection. While effective, existing DA methods suffer from a substantial drop in performance when applied to a multi-class training setting, due to the co-existence of low-quality pseudo labels and class imbalance issues. In this paper, we address this challenge by proposing a novel \textbf{ReDB} framework tailored for learning to detect all classes at once. Our approach produces \textbf{R}eliable, \textbf{D}iverse, and class-\textbf{B}alanced pseudo 3D boxes to iteratively guide the self-training on a distributionally different target domain. To alleviate disruptions caused by the environmental discrepancy (e.g., beam numbers), the proposed cross-domain examination (CDE) assesses the correctness of pseudo labels by copy-pasting target instances into a source environment and measuring the prediction consistency. To reduce computational overhead and mitigate the object shift (e.g., scales and point densities), we design an overlapped boxes counting (OBC) metric that allows to uniformly downsample pseudo-labeled objects across different geometric characteristics. To confront the issue of inter-class imbalance, we progressively augment the target point clouds with a class-balanced set of pseudo-labeled target instances and source objects, which boosts recognition accuracies on both frequently appearing and rare classes. Experimental results on three benchmark datasets using both voxel-based (\textit{i.e.}, SECOND) and point-based 3D detectors (\textit{i.e.}, PointRCNN) demonstrate that our proposed ReDB approach outperforms existing 3D domain adaptation methods by a large margin, improving 23.15\% mAP on the nuScenes $\rightarrow$ KITTI task. The code is available at \href{https://github.com/zhuoxiao-chen/ReDB-DA-3Ddet}{https://github.com/zhuoxiao-chen/ReDB-DA-3Ddet}.

\end{abstract}

\vspace{-2ex}
\section{Introduction}

\begin{figure}
\centering
\begin{subfigure}{0.9\linewidth}
\centering
   \includegraphics[width=1\textwidth]{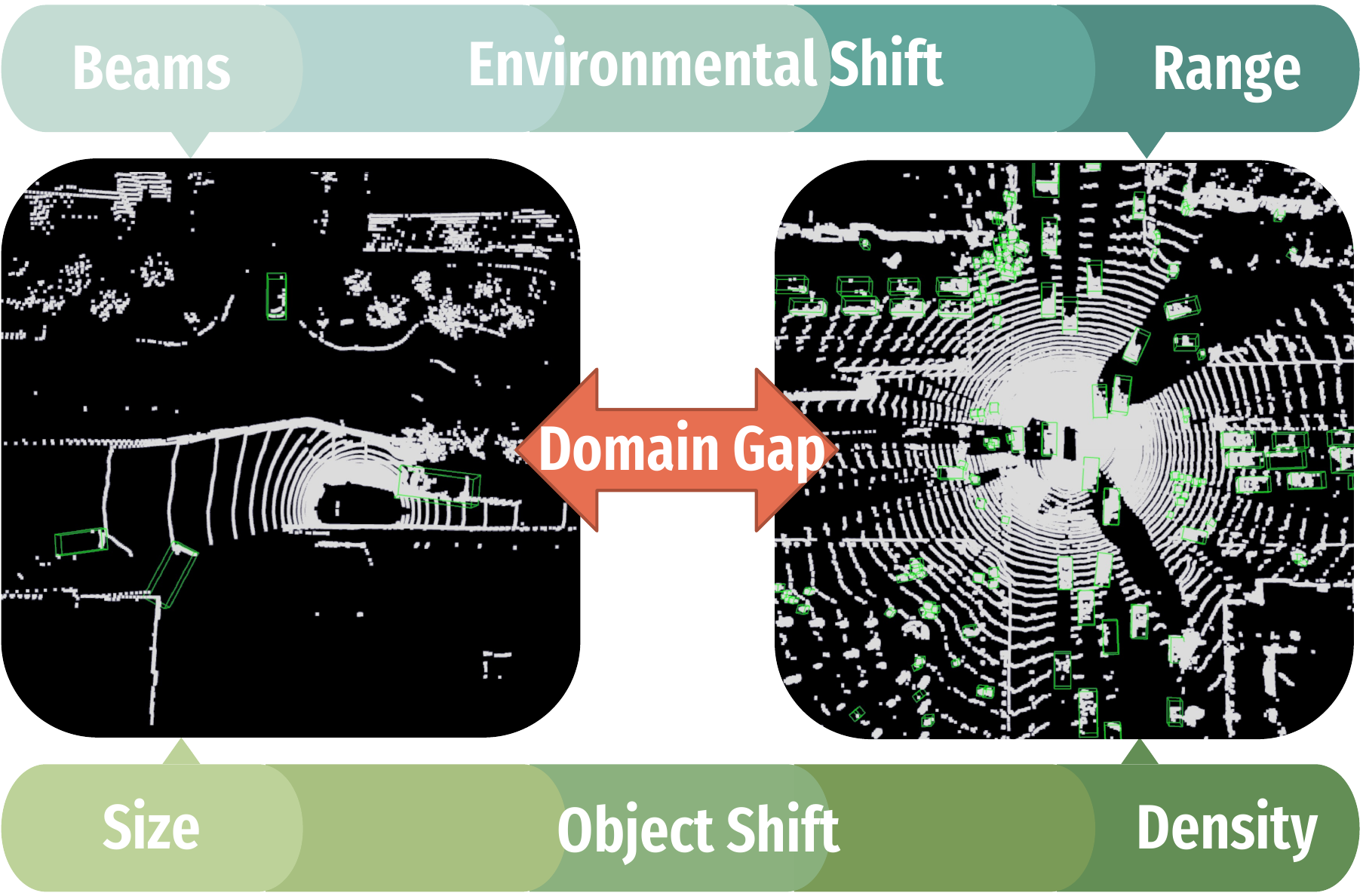}
\end{subfigure}
\begin{subfigure}{0.9\linewidth}
   \includegraphics[width=1\textwidth]{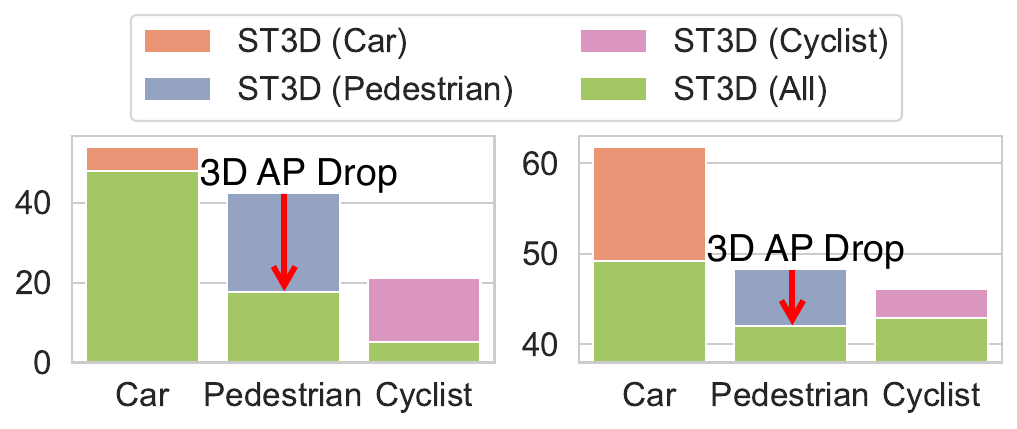}
\end{subfigure}
\caption[]{\textit{Top}: An illustration of the domain gap in 3D point clouds. \textit{Bottom}: Average Precision (AP) drop of ST3D \cite{DBLP:conf/cvpr/YangS00Q21} when applied to the multi-class adaptation from nuScenes to KITTI (\textit{left}), and from Waymo to  KITTI (\textit{right}).\vspace{-2ex} \label{fig:brief_framework}

}
\end{figure}




As LiDAR-based 3D object detection continues to gain traction in various applications such as robotic systems \cite{DBLP:conf/iros/AhmedTCMW18, DBLP:conf/iros/MontesLCD20, DBLP:journals/sensors/WangLSLZSQT19, DBLP:conf/cvpr/TremblayTB18, DBLP:journals/rcim/ZhouLFDMZ22, 8024025} and self-driving automobiles \cite{DBLP:conf/nips/DengQNFZA21, DBLP:conf/eccv/WangLGD20, DBLP:journals/pr/QianLL22a, DBLP:conf/cvpr/WangCGHCW19, DBLP:journals/tits/ArnoldADFOM19, DBLP:conf/iclr/YouWCGPHCW20, DBLP:journals/corr/abs-2206-09474, DBLP:conf/cvpr/MeyerLKVW19, DBLP:conf/iclr/LuoCWYHB23, DBLP:conf/cvpr/LiOSZW19, DBLP:conf/icip/McCraeZ20}, it becomes increasingly vital to address the challenges of deploying detectors in real-world scenes. The primary obstacles stem from the discrepancy between the training and test point cloud data, which are commonly curated from different scenes, locations, times, and sensor types, creating a \textit{domain gap}.
This domain gap mainly comes from the \textit{object shift} and \textit{environmental shift}, and can significantly degrade the prediction accuracy of 3D detectors. Object shift \cite{DBLP:conf/cvpr/WangCYLHCWC20, DBLP:conf/iccv/LuoCZZZYL0Z021} refers to the changes in the spatial distribution, point density, and scale of objects between the training and test domains. For instance, the average length of cars in the Waymo dataset \cite{DBLP:conf/cvpr/SunKDCPTGZCCVHN20} is significantly different from the one in the KITTI dataset \cite{DBLP:conf/cvpr/GeigerLU12} by around 0.91 meters \cite{DBLP:conf/cvpr/WangCYLHCWC20}. Environmental shift, on the other hand, arises from composite differences in the surrounding environment such as inconsistent beam numbers, angles, point cloud ranges, and data acquisition locations. For example, in Fig \ref{fig:brief_framework}, Waymo generates 3D scenes by 64-beam LiDAR sensors, while nuScenes \cite{DBLP:conf/cvpr/CaesarBLVLXKPBB20} comprises more sparse 32-beam environments with double large beam angles.

The co-existence of object shift and environmental shift poses a great challenge to the 3D detection models to be deployed in the wild. To combat such a problem, domain-adaptive 3D detection approaches \cite{DBLP:conf/iccv/LuoCZZZYL0Z021, yang2022st3d++, DBLP:conf/cvpr/YangS00Q21, DBLP:conf/nips/ZengWWXYYM21, DBLP:conf/cvpr/Zhang0021} have been exploited to adapt the model trained on a labeled dataset (\textit{i.e.}, source domain) to an unlabeled dataset from a different distribution (\textit{i.e.}, target domain). In this area, pioneering research of ST3D \cite{DBLP:conf/cvpr/YangS00Q21} presents a self-training paradigm that generates pseudo-labeled bounding boxes to supervise the subsequent learning on the target point clouds. In the same vein, later studies \cite{DBLP:conf/iccv/LuoCZZZYL0Z021, DBLP:conf/nips/ZengWWXYYM21, yang2022st3d++, DBLP:conf/cvpr/YouLPCSHCW22, DBLP:conf/nips/YouPLZCH0W22} seek different solutions based on self-supervised techniques, such as mean-teacher framework \cite{DBLP:conf/iccv/LuoCZZZYL0Z021}  and contrastive learning \cite{DBLP:conf/nips/ZengWWXYYM21} for stable optimization and obtaining better embeddings.

\noindent \textit{Revisiting Domain-adaptive 3D Detection Setup.}  The aforementioned domain-adaptive 3D detection approaches have typically followed a \textbf{single-class training} setting, where the models are trained to adapt to each class separately.  While it is more practical and fairer to train the models with all classes, our empirical study has shown that the detection performance of prior works decreases considerably when switching to a multi-class setting (Fig \ref{fig:brief_framework}). Such an average precision (AP) drop can be attributed to the poor quality of produced pseudo labels (\textit{i.e.}, erroneous and redundant) and the lower recognition accuracy of rare classes (\textit{e.g.}, 91 times fewer bicycles than cars in Waymo \cite{DBLP:conf/cvpr/SunKDCPTGZCCVHN20}). 

\begin{figure*}[t]
    \centering
    \includegraphics[width=1\linewidth]{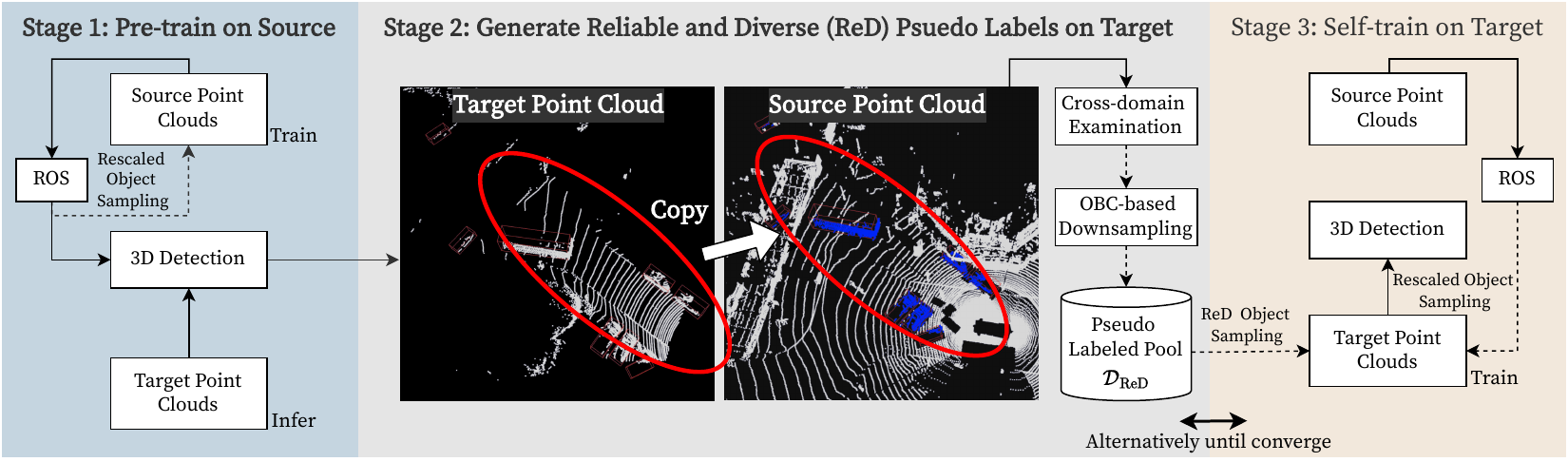}
    \caption{The overall framework of the proposed method: we firstly obtain high confident pseudo labels from Stage 1. In Stage 2, we check the reliability of pseudo labels by the cross-domain examination (CDE) and reduce geometrical redundancy by OBC-based downsampling. Finally, we progressively augment target point clouds by injecting \textsc{ReD}  labels and source labels for class-balanced self-training in Stage 3. \eat{, comprising three stages. In the Stage 1, the 3D detector is pre-trained with randomly scaled source data. The Stage 2 involves generating reliable and diverse (\textsc{ReD} )  pseudo labels via the proposed CDE-based quality control module and OBC-based diversity module. In the Stage 3, we train the 3D detector with class-balanced distributed \textsc{ReD}  labels and source aided labels. Finally, we alternatively generate target pseudo labels and self-train the target data until the 3D detector is optimized.}\vspace{-1ex}}
    \label{fig:bframework}
\end{figure*}

In this work, we propose a novel \textsc{ReDB} framework (Fig \ref{fig:bframework}) that aims to generate \underline{Re}liable, \underline{D}iverse, class-\underline{B}alanced pseudo labels for the domain-adaptive 3D detection task. Our approach addresses the challenge of multi-class learning by incorporating the following three mechanisms:

\noindent\textbf{Reliability: Cross-domain Examination.} To remove erroneous pseudo labels of high confidence and avoid error accumulation in self-training,
we introduce a cross-domain examination (CDE) strategy to assess pseudo label reliability. After copying the pseudo-labeled target objects into a familiar source environment, the reliability is measured by the consistency \textit{i.e.}, Intersection-over-Union (IoU) between two box predictions in the target domain and the source domain, respectively. Any objects with low IoUs will be considered unreliable and then discarded. To prevent point conflicts between the source and target point clouds, we remove the source points that fall within the region where pseudo-labeled objects will be copied to. The proposed CDE ensures that the accepted pseudo-labeled instances are domain-agnostic and less affected by environmental shifts. 


\noindent\textbf{Diversity: OBC-based Downsampling.} To avoid redundant pseudo labels that are frequently and similarly scaled, it is important to prevent the trained detector from collapsing into a fixed pattern that may only detect certain types of objects (\textit{e.g.}, small-sized cars), and miss other instances of unique styles (\textit{e.g.}, buses and trucks). In order to enhance geometric diversity, we derive a metric called overlapped boxes counting (OBC) metric to uniformly downsample the pseudo labels. The metric design is motivated by the observation that 3D detector tends to predict more boxes for objects with uncommon geometries, as they are harder to localize with only a few tight boxes. We count the number of regressed boxes surrounding each detected object as OBC and use kernel density estimation (KDE) to estimate its empirical distribution. We then apply downsampling according to the inverse probability of KDE, effectively reducing the number of pseudo labels in the high-density OBC region, where objects have similar and frequent geometries. By learning from a diverse subset of pseudo labels, the 3D detector can better identify objects of various scales and point densities, potentially mitigating object shift.

\noindent\textbf{Balance:  Class-balanced Self-Training.} Despite the fact that reliable and diverse pseudo labels can be selected by the previous two modules, there still exists a significant inter-class imbalance. To achieve class-balanced self-training, we randomly inject pseudo-labeled objects into each target point cloud, with an equal number of samples from each category. By learning from such class-balanced target data, the model can better grasp the holistic semantics of the target labels. To enable a smooth transition of learning from the source data to the target data, we first augment the target data with labeled source objects in a class-balanced manner for the initial few steps. We then gradually reduce the ratio of source objects and increase the number of \textsc{ReD} labels as the self-training proceeds. This progressive class-balanced self-training allows the model to adapt stably to the target domain, enhancing recognition for both frequently appearing and rare classes. 

Our work rectifies the setup of domain-adaptive 3D detection to a multi-class scenario and proposes a novel \textsc{ReDB} framework for pseudo labeling in  domain-adaptive 3D detection. Extensive experiments on three large-scale testbeds evidence that the proposed \textsc{ReDB} is of exceptional adaptability for both voxel-based and point-based contemporary 3D detectors in varying environments, improving 3D mAP of 20.66\% and 23.15\% over state-of-the-art methods on the nuScenes $\rightarrow$ KITTI tasks, respectively. \eat{Note that for a fair evaluation of our proposed \textsc{ReDB}, we rerun the existing approaches in a multi-class setting.}

\section{Related Work}

\noindent \textbf{3D Object Detection from Point Clouds.} With the growing advancement of neural networks over the last decade, the 3D point clouds can be effectively encoded by deep models to either point-based representation or grid-based representation. The series of point-based detectors \cite{DBLP:conf/cvpr/ShiWL19, DBLP:conf/iccv/YangS0SJ19, DBLP:conf/cvpr/YangS0J20, DBLP:conf/cvpr/ShiR20, DBLP:conf/cvpr/PanXSLH21, DBLP:conf/cvpr/WangS0FQ0S022, DBLP:conf/cvpr/ZhangHXMWG22} directly encode 3D objects from raw points using PointNet encoder, then generate 3D proposals from point features. This type of encoding strategy can aid the model better preserve and learn the geometric properties of 3D objects. While, the grid-based detectors \cite{DBLP:conf/icra/EngelckeRWTP17, DBLP:conf/iros/Li17, DBLP:conf/cvpr/YangLU18, DBLP:conf/cvpr/ZhouT18, DBLP:journals/sensors/YanML18, DBLP:conf/cvpr/LangVCZYB19, DBLP:conf/eccv/WangFKRPFS20, DBLP:journals/pami/ShiWSWL21, DBLP:conf/aaai/DengSLZZL21, DBLP:conf/cvpr/YinZK21, DBLP:conf/eccv/ShiLM22, DBLP:conf/eccv/SunT0LXLA22, DBLP:conf/cvpr/FanPZWZWWZ22} rely on convolution neural networks for efficiently capturing regularly voxelized point clouds. Hybrid methods \cite{DBLP:conf/cvpr/HeLLZ22, DBLP:conf/cvpr/HuKW22, yang2022st3d++, DBLP:conf/cvpr/YangS00Q21} take advantages of both paradigms, yielding better results at the expense of processing speed. However, the preceding work overlooks the domain gap in different 3D scenes such that could scarcely be applied to unseen scenarios. 

\noindent \textbf{Domain Adaptation for 3D Detection.} To surmount such difficulty, unsupervised domain adaptation (UDA) strives to transfer knowledge from a labelled source domain to an unlabeled target domain. There are primarily two types of strategies to eliminate the discrepancy between two domains. One is the adversarial-based approach \cite{DBLP:conf/icml/GaninL15, DBLP:journals/jmlr/GaninUAGLLML16, DBLP:conf/cvpr/TzengHSD17, DBLP:conf/cvpr/SaitoWUH18, DBLP:conf/icml/HoffmanTPZISED18, DBLP:conf/iccv/DengLZ19, 10.1145/3469877.3490600, 10107906, DBLP:conf/icml/LuoWHB20, DBLP:conf/mm/LuoHW0B20, DBLP:conf/iccv/00010LXYKFW21, DBLP:conf/iccv/0014LZ021} that learns the domain-invariant features. The other is the statistical-based approach \cite{DBLP:conf/icml/LongZ0J17, DBLP:conf/nips/LongZ0J16, DBLP:conf/icml/LongC0J15, DBLP:conf/eccv/SunS16, DBLP:conf/aaai/ChenFCJCJ020, DBLP:conf/mm/WangLHB20, DBLP:conf/icmcs/WangLZ0H22, DBLP:journals/pami/LiXLLHW22, DBLP:journals/pami/LongCCWJ19} that seeks to minimize the distribution distance between two domains. Despite the fact that UDA techniques have been extensively investigated in the last decade, few UDA approaches aim to address the domain shift existing in 3D object detection. The distribution shifts of 3D detection are identified as dissimilar object statistics \cite{DBLP:conf/cvpr/WangCYLHCWC20}, weather effects \cite{DBLP:conf/iccv/XuZ0QA21, DBLP:conf/cvpr/HahnerSBHYDG22}, sensor difference \cite{iv_rist2019cross, DBLP:conf/iccv/Gu0XFCLSM21, DBLP:conf/eccv/WeiWRLZL22} or synthesis-to-real gap \cite{DBLP:conf/iccvw/SalehAAINHN19, DBLP:journals/ral/DeBortoliLKH21, DBLP:conf/cvpr/LehnerGMSMNBT22}. To close the domain shift, an adversarial-based approach \cite{DBLP:conf/cvpr/Zhang0021} was proposed to match features of different scales and ranges. Another streams of work focus on generating and enhancing 3D pseudo labels through memory bank \cite{DBLP:conf/cvpr/YangS00Q21, yang2022st3d++}, mean-teacher paradigm \cite{DBLP:conf/iccv/LuoCZZZYL0Z021} or contrastive learning \cite{DBLP:conf/nips/ZengWWXYYM21}. However, existing methods necessitate either extra computational time or significant memory usage. Besides that, these methods generate low-quality pseudo-labels induced by environmental gaps and numerous analogous geometrical features, and overlook the class-imbalanced issue. 

\section{Methodology}
\subsection{Problem Definition}
In this section, we mathematically formulate the problem of unsupervised domain adaptation for 3D object detection and set up the notations. The main objective is to adapt a 3D object detector trained on the labeled point clouds $\{(X^s_i, Y^s_i)\}^{N_s}_{i=1}$ from the source domain, to the unlabeled data $\{X^t_i\}^{N_t}_{i=1}$ in the target domain. $N_s$ and $N_t$ indicate the number of point clouds in the source and target domains, respectively. The 3D box annotation of the $i$-th source point cloud is denoted as $Y^s_i= \{b_j\}^{B_i}_{j=1}$ and $b_j = (x, y, z, w, l, h, \theta, c)\in\mathbb{R}^8$, where $B_i$ is the total number of labeled boxes in $X^s_i$, $(x, y, z)$ represents the box center location, $(w, l, h)$ is the box dimension, $\theta$ is the box orientation, and $c \in \{1, ..., C\}$ is the object category. 


\subsection{Overall Framework}
\begin{figure}[t]
    \centering
    \includegraphics[width=1\linewidth]{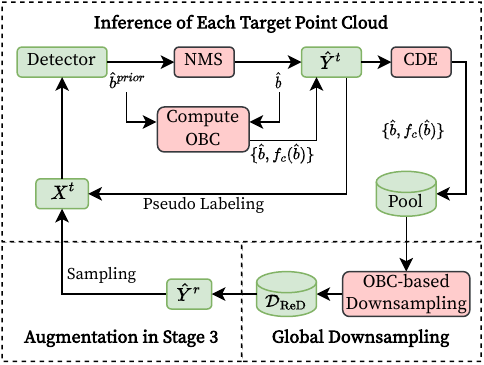}
    \caption{In Stage 2 of our approach, we input each target point cloud $X^t$ to the pre-trained detector and obtain 3D box predictions. By applying NMS, we select confident and non-overlapped boxes $\hat{b}$ as pseudo labels. Simultaneously, we compute OBC score for each pseudo-labeled $\hat{b}$, based on prior-NMS boxes $\hat{b}^{prior}$. Next, we use CDE to add qualified $\hat{b}$ to a pool. After inferring all target data, we globally downsample the pseudo-labeled boxes in the pool using OBC, resulting in $\mathcal{D}_{\operatorname{ReD}}$. In Stage 3, we augment each target point cloud by sampling high-quality objects from $\mathcal{D}_{\operatorname{ReD}}$ (Eq \ref{eq:6}) in a class-balanced manner during self-training. Stage 2 and 3 are alternated until convergence. \vspace{-2ex}}
    \label{fig:rebuttal_framework}
\end{figure}

Our approach consists of three stages to facilitate knowledge transfer from the label-rich source domain to the label-scarce target domain, as illustrated in Fig \ref{fig:bframework}. In Stage 1, the 3D detector (\textit{e.g.}, SECOND \cite{DBLP:journals/sensors/YanML18} or PointRCNN \cite{DBLP:conf/cvpr/ShiWL19}) is pre-trained on the source domain using the standard random object scaling (ROS) augmentation \cite{DBLP:conf/cvpr/YangS00Q21}. Upon reaching convergence of pre-training, the unlabeled point clouds are passed to the pre-trained detector to generate pseudo-labels $\{\widehat{Y}^t_i\}^{N_t}_{i=1}$ of high confidence for the target data $\{X^t_i\}^{N_t}_{i=1}$. Specifically, the produced pseudo labels will undergo scrutiny by the cross-domain examination (CDE) strategy (Sec \ref{sec:cde}) and down-sampled by the OBC-based diversity module  (Sec \ref{sec:obc}), forming a subset of reliable and diverse (\textsc{ReD}) objects associated with pseudo labels.
The details of stage 2 are illustrated in Fig \ref{fig:rebuttal_framework}. During model self-training on the target set in Stage 3, we randomly inject \textsc{ReD} target objects and source objects into each target point clouds in a class-balanced manner (Sec \ref{sec:balance}), with the ratio of source samples gradually decreasing. The 3D detector is iteratively trained to adapt by alternating between Stage 2 and Stage 3.

\begin{figure}[t]
    \centering
    \includegraphics[width=1\linewidth]{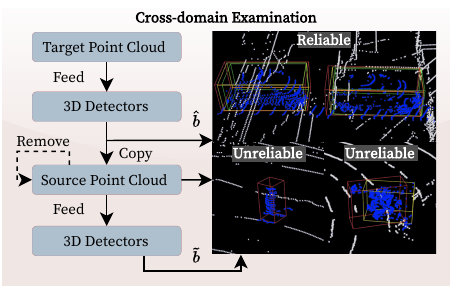}
    \caption{The proposed cross-domain examination (CDE) strategy involves copying pseudo target points $X^{ps}$ in blue, with red and yellow boxes representing initial predictions in the target and source environments, respectively. Green boxes denote the ground truth for reference. If the intersection over union (IoU) between the target and source predictions is large, like in the first example, the reliability is accepted. However, in cases where $\hat{b}$ is not detected or the IoU is small, like in the second and third examples, these labels are considered unreliable due to inconsistencies caused by environmental shift.\vspace{-2ex}}
    \label{fig:cde}
\end{figure}

\subsection{Reliability: Cross-domain Examination}\label{sec:cde}

The proposed cross-domain examination (CDE) strategy aims to identify high-quality pseudo labels that are robust to the environmental shift. First, given a 3D detector $F(\,\cdot\,; \theta^s)$ pre-trained on the source domain, we can obtain the initial pseudo labels by inferring a target point cloud $X_i^t$ as:
\begin{equation}
    \begin{aligned}
    \widehat{Y}_i^t = \{\hat{b}_j\}^{\widehat{B}_i}_{j=1} = F(X_i^t; \theta^s),
    \end{aligned}
\end{equation}
where $\theta^s$ is the pre-trained weights, and $\widehat{Y}^t\in\mathbb{R}^{\widehat{B}\times 8}$ denotes the pseudo labels with the model confidence greater than $\delta_{\operatorname{pos}}$. $\widehat{B}_i$ is the number of obtained pseudo labels. Below, we omit the subscript $i$ for simplicity. To attain a real source environment, we randomly sample a source point cloud $X^s \sim rand(\{(X^s_i)\}^{N_s}_{i=1})$ and copy-paste the target point clouds $X^{ps}\subset X^t$ that fall within $\widehat{Y}^t$, into the sampled source point cloud as shown in Fig \ref{fig:cde}. To prevent conflicts between existing points and the pseudo points to be pasted, we apply the object points removal operator $\texttt{R}(\,\cdot\,)$ to remove source points residing in the area overlapped with the copied target samples $X^{ps}$. Now, we can safely move the target points to the source point clouds as $\texttt{R}(X^s) \oplus X^{ps}$, where $\oplus$ represents point concatenation. We generate new predictions $\widetilde{Y}^t$ of high confidence for the concatenated source point clouds as:
\begin{equation}
    \begin{aligned}
     \widetilde{Y}_i^t = \{\tilde{b}_j\}^{\widetilde{B}_i}_{j=1} = F(\texttt{R}(X^s)\oplus X^{ps}); \theta^s),
    \end{aligned}
\end{equation}
where $\widetilde{B}_i$ represents the number of boxes in the updated source clouds. To examine the resiliency of the pseudo-labeled object $\hat{b}\in\hat{Y}^t$ to the environmental shift, we compute the intersection over union (IoU) between the initial predicted box $\hat{b}$ and its corresponding prediction $\tilde{b}\in\tilde{Y}^t$ in the source environment. Through comparing the IoU to a predefined IoU threshold $\delta_{\operatorname{cde}}$, we can effectively discriminate reliable pseudo labels from unreliable ones by using the following criterion: 
\begin{equation}\label{eq:cde}
    \begin{aligned}
    f_{\operatorname{CDE}}(\hat{b}; \tilde{b})=
    \begin{cases}
     \hat{b} & \text{IoU}(\hat{b}, \tilde{b}) \geq \delta_{\operatorname{cde}}, \\
     \emptyset & \text{otherwise}.
    \end{cases}
    \end{aligned}
\end{equation}
We apply the CDE only at the first round of pseudo labelling because the pre-trained detector has no knowledge about target environments. The process of proposed CDE strategy is presented in Fig \ref{fig:cde} with three examples for illustration.

\subsection{Diversity: OBC-based Downsampling}\label{sec:obc}

\begin{figure}[t]
    \centering
    \includegraphics[width=1\linewidth]{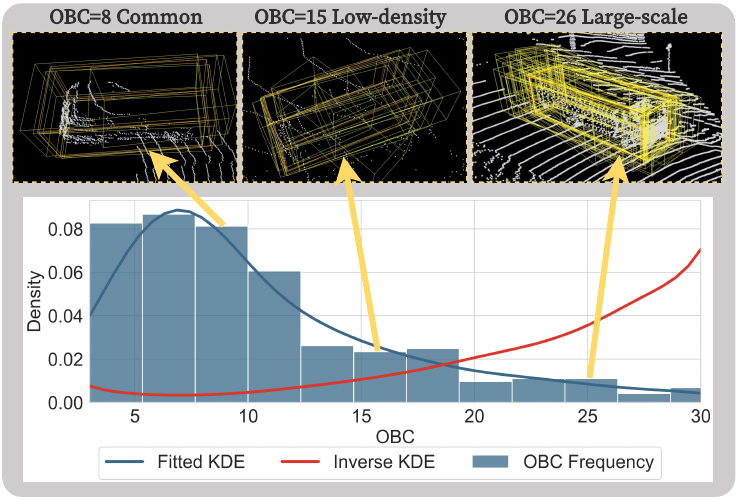}
    \caption{An illustration of overlapped boxes counting (OBC). The upper part shows box predictions before NMS generated around three positively predicted objects associated with different OBC values. The bottom part is the plot demonstrating the distributions of OBC values of all detected objects, with the fitted KDE (blue) and inverse KDE (red) for diverse sampling.\vspace{-1.5ex}}
    \label{fig:OBC}
\end{figure}

Although we have already removed unreliable pseudo labels, the remaining label set may contain objects with similar patterns. This hinders the model from learning a more comprehensive distribution of the target domain. To ensure that the pseudo label set is representative of the broad spectrum of geometric characteristics present in target data, we investigate ways to properly measure object diversity. Our empirical study shows that 3D detectors generate more box predictions around unfamiliar objects, as visualized in Fig \ref{fig:OBC}. It is observed that, the box predictions (in yellow) before non-maximum suppression (NMS) have obvious overlaps (\textit{i.e.}, $\text{IoU}>\delta_{\operatorname{obc}}$) with the corresponding final predicted box (in red). Besides, predicted boxes for low-density objects typically orientate differently, while the regressed box scales for large objects have a high variance.

This observation motivates us to derive the concept of overlapped boxes counting (OBC), which counts the number of predicted boxes that surround the same object with the $\text{IoU} >\delta_{\operatorname{obc}}$. The OBC can comprehensively reflect the uniqueness in geometric characteristics of objects, including scales, locations, point densities. A qualitative analysis of OBC is provided in Sec \ref{sec:vis_OBC}, which evidence that pseudo-labeled target objects with high OBC values tend to be uncommon in different geometric aspects. To estimate the probability density function (PDF) of OBC values, we use bar plots to show the frequency of different OBC values, and fit it with Kernel Density Estimation (KDE) (as shown in the blue curve of Fig \ref{fig:OBC}). It can be seen from the plot that the distribution of OBC is very skewed, implying that the majority of geometric features are quite analogous since fell at high frequency interval (\textit{i.e.}, from 3 to 12). Conversely, objects with rare geometrics fell on the long distribution tail (\textit{i.e.}, from 12 to 30). To get a diverse subset and exclude too many objects in high density regions, we propose to leverage the inverse KDE function (shown in red curve) to assign a low sampling weight for pseudo-labeled objects fell in dense regions, while high sampling probability for objects in tail regions. In particular, we firstly collect the set $\mathcal{O}$ of all OBC values with the cardinality $\hat{B} = \sum_{i=1}^{N_t} \hat{B}_i$:
\begin{equation}
    \mathcal{O} = \{f_c(\hat{b}_j) | b_j\in\hat{Y}_i^t\}_{i=1}^{N_t},\vspace{-1ex}
\end{equation}
where $f_c(\,\cdot\,)$ is the function to count OBC value for each pseudo-labeled box $\hat{b}$. The KDE with the Gaussian kernel of the random variable $O$ can be calculated with a finite set of $\mathcal{O}$:
\begin{equation}
     f_{\operatorname{KDE}}(O)=\frac{1}{\hat{B} \sigma \sqrt{2 \pi}} \sum_{j=1}^{\hat{B}} e^{-\frac{1}{2}(\frac{f_c(\hat{b}_j)-O}{\sigma})^2},
\end{equation}
where $\sigma$ is the standard deviation. The sampled subset $\mathcal{D}_{\operatorname{ReD}}$ of size $\hat{B}/d$ are uniformly downsampled from the inverse KDE, where $d>1$ is the down sampling rate. Finally, the sampled subset $\mathcal{D}_{\operatorname{ReD}}$ contains pseudo-labelled objects with more uniformly distributed OBC values, indicating less common patterns and more diverse geometrics (\textit{e.g.}, object scales, distances and densities). 


\subsection{Balance: Class-balanced Self-Training}\label{sec:balance}
Regardless OBC-based downsampling creates a pool of reliable and intra-class diverse (\textsc{ReD}) pseudo-labels, but the inter-class imbalance issue still poses a challenge for multi-class adaptation.  We aim to leverage a class-balanced paradigm to inject the generated pseudo labels to each target point cloud, during the self training stage:
\begin{equation} \label{eq:6}
    \begin{aligned}
    \widehat{Y}^{r} &= \{\{\hat{b}_j^1\}^{S^{r}}_{j=1} \oplus \cdots \oplus\{\hat{b}_j^C\}^{S^{r}}_{j=1}\}  \sim \mathcal{D}_{\operatorname{ReD}}, \\
    \end{aligned}
\end{equation}
where $\oplus$ represents the point cloud concatenation and the superscript $1,\cdots, C$ of $\hat{b}$ indicates the predicted class of the pseudo label, and $S^{r}$ is the sampling number of target \textsc{ReD} labels for each class. To moderate and stable the adaptation process with the huge domain gap, we augment the target data by injecting source ground truth (GT) labels with ROS augmentation: 
\begin{equation}
    \begin{aligned}
    P^{g} &= \{Y_1^s \oplus \dots \oplus Y_{N_s}^s \}, \\
    Y^{g} &= \{\{b_j^1\}^{S^{g}}_{j=1} \oplus \cdots \oplus\{b_j^C\}^{S^{g}}_{j=1}\}  \sim P^{g}, \\
    \end{aligned}
\end{equation}
where $S^{g}$ is the GT sampling number for each class, $P^{g}$ is the source GT pool containing labelled boxes from all source point clouds. We gradually \textit{increase} $S^{r}$ and \textit{reduce} $S^{g}$ during the self training, that enables 3D detector to smoothly generalize to the target domain. Finally, we concatenate the sampled target pseudo-labeled \textsc{ReD} boxes and source GT boxes to the initial pseudo labels, including the points $X^{r}$ and $X^{g}$ inside target \textsc{ReD} boxes and source GT boxes, respectively:
\begin{equation}
    \begin{aligned}
    \widehat{Y}^t&\leftarrow\widehat{Y}^t \oplus \widehat{Y}^{r} \oplus Y^{g}, \\
    X^t&\leftarrow X^t \oplus X^{r} \oplus X^{g}.
    \end{aligned}
\end{equation}
Finally, we have each target point cloud augmented with class-balanced \textsc{ReDB} pseudo labels as $(X_i^t, \widehat{Y}_i^t)$, and train the detector as: 
\begin{equation}
    \begin{aligned}
     F(\, \cdot \, ;\theta^t) \leftarrow \{(X^t_i, \widehat{Y}^t_i)\}^{N_t}_{i=1}.
    \end{aligned}
\end{equation}
We alternatively generate target pseudo labels (Stage 2) and self-train the model (Stage 3),  until the 3D detector fully is optimized, as illustrated in Fig \ref{fig:bframework}. The detailed Algorithm is summarized in the supplementary materials.


\begin{table*}[htbp]
\renewcommand\arraystretch{1.15}
    \centering
    \caption{ Experiment results of three adaptation tasks are presented, with the reported average precision (AP) for bird’s-eye view ($\text{AP}_{\text{BEV}}$) / 3D ($\text{AP}_{\text{3D}}$) of car, pedestrian, and cyclist with IoU threshold set to 0.7, 0.5, and 0.5 respectively. When the KITTI is the target domain, we report the AP at \textbf{Moderate} difficulty. The last column shows the mean AP for all classes. We indicate the best adaptation result by \textbf{bold} and we highlight the row representing our method.}
    \begin{small}
    \setlength{\tabcolsep}{5.1mm}{
        \begin{tabular}{c|c|c|c|c|c}
            \bottomrule[1pt]
            \textsc{Task} & \textsc{Method}  & \textsc{Car} & \textsc{Pedestrian} & \textsc{Cyclist} & \textsc{Mean AP} \\
            \hline
            \multirow{6}{*}{Waymo $\rightarrow$ KITTI} & \textsc{Source Only}  & 51.48 / 19.78 & 40.80 / 31.26 & 47.63 / 35.45 & 46.64 / 28.83 \\
            
            & SN & 76.61 / \textbf{54.14} & \textbf{52.48} / \textbf{48.20} & 34.56 / 32.74 & 54.55 / 45.03 \\
            & ST3D & 77.62 / 49.24 & 44.45 / 42.04 & 47.74 / 42.95 & 56.60 / 44.70 \\
            & ST3D++ & 77.68 / 50.03 & 49.09 / 46.19 & 51.50 / 47.70 & 59.42 / 47.97 \\
            
            & \textbf{\textsc{ReDB}} \cellcolor{LightCyan} &  \textbf{80.37}\cellcolor{LightCyan} / 54.12\cellcolor{LightCyan} & 51.01\cellcolor{LightCyan} / \textbf{48.20}\cellcolor{LightCyan} & \textbf{52.05}\cellcolor{LightCyan} / \textbf{47.97}\cellcolor{LightCyan} & \textbf{61.14}\cellcolor{LightCyan} / \textbf{50.10}\cellcolor{LightCyan}  \\
            \cline{2-6}
            & \textsc{Oracle} &  83.29 / 73.45 & 46.64 / 41.33 & 62.92 / 60.32 & 64.28 / 58.37  \\

            \toprule[1pt]
            \bottomrule[1pt]
            \multirow{6}{*}{Waymo $\rightarrow$ nuScenes} & \textsc{Source Only}  & \textbf{30.64} / 17.18 & 1.81 / 0.58 & 0.97 / 0.88 & 11.14 / 6.22 \\
            & SN & 29.56 / 17.78 & 1.33 / 1.07 & 2.92 / 2.61 & 11.27 / 7.15 \\
            & ST3D & 28.42 / 17.83 & 1.64 / 1.39 & 4.01 / 3.54 & 11.36 / 7.59 \\
            & ST3D++ & 28.87 / \textbf{19.15} & 1.82 / 1.58 & 4.09 / 3.74 & 11.60 / 8.16 \\
            & \textbf{\textsc{ReDB}}\cellcolor{LightCyan} & 30.12\cellcolor{LightCyan} / 18.56\cellcolor{LightCyan} & \textbf{2.47}\cellcolor{LightCyan} / \textbf{2.14}\cellcolor{LightCyan} & \textbf{6.56}\cellcolor{LightCyan} / \textbf{5.19} &  \textbf{13.05}\cellcolor{LightCyan} / \textbf{8.63}\cellcolor{LightCyan} \\
            \cline{2-6}
            & \textsc{Oracle} &  51.88 / 34.87 & 25.24 / 18.92 & 15.06 / 11.73 & 30.73 / 21.84 \\
            \toprule[1pt]
            \bottomrule[1pt]

            \multirow{6}{*}{nuScenes $\rightarrow$ KITTI} 
            
            & \textsc{Source Only} & 39.15 / 7.65 & 21.54 / 16.87 & 6.31 / 2.44 & 22.33 / 8.98 \\
            
            & SN & 56.08 / 28.67 & 23.05 / 16.84 & 5.11 / 2.32 & 28.08 / 15.94 \\

            & ST3D & 71.50 / 48.09 & 22.64 / 17.61 &  7.86 / 5.20 & 34.00 / 23.64 \\
            
            & ST3D++ & 69.90 / 44.62 & 24.11 / 18.20 &  10.14 / 6.39 & 33.75 / 21.64 \\

            
            & \textbf{\textsc{ReDB}} \cellcolor{LightCyan} & \textbf{74.23}\cellcolor{LightCyan} / \textbf{51.31}\cellcolor{LightCyan} & \textbf{25.95}\cellcolor{LightCyan} / \textbf{18.38}\cellcolor{LightCyan} & \textbf{13.82}\cellcolor{LightCyan} / \textbf{8.64}\cellcolor{LightCyan} & \textbf{38.00}\cellcolor{LightCyan} / \textbf{26.11}\cellcolor{LightCyan}  \\
            \cline{2-6}
            & \textsc{Oracle} & 83.29 / 73.45 & 46.64 / 41.33 & 62.92 / 60.32 & 64.28 / 58.37 \\
            \toprule[0.8pt]
        \end{tabular}
    }
    \end{small}
    \label{tab:SOTAcomparison}

\end{table*}

\begin{table*}[htbp]
\renewcommand\arraystretch{1.15}
    \centering
    \caption{ Experiment results at \textbf{Hard} difficulty of adapting to the KITTI dataset. We report average precision (AP) for bird’s-eye view ($\text{AP}_{\text{BEV}}$) / 3D ($\text{AP}_{\text{3D}}$) of car, pedestrian, and cyclist with IoU threshold set to 0.7, 0.5, and 0.5 respectively. The last column shows the mean AP for all classes. We indicate the best adaptation result by \textbf{bold}, and we highlight the row representing our method.}
    \begin{small}
    \setlength{\tabcolsep}{5.1mm}{
        \begin{tabular}{c|c|c|c|c|c}
            \bottomrule[1pt]
            \textsc{Task} & \textsc{Method}  & \textsc{Car} & \textsc{Pedestrian} & \textsc{Cyclist} & \textsc{Mean AP} \\
            \hline
            \multirow{4}{*}{Waymo $\rightarrow$ KITTI}
            & SN & 75.57 / \textbf{53.87} & \textbf{49.15} / \textbf{45.58} & 33.09 / 31.19 & 52.60 / 43.55 \\
            & ST3D & 75.31 / 48.01 & 41.42 / 43.02 & 45.76 / 40.76 & 54.16 / 43.93 \\
            & ST3D++ & 75.54 / 48.03 & 45.63 / 42.00 & 47.12 / 43.70 & 56.10 / 44.58 \\
            
            & \textbf{\textsc{ReDB}} \cellcolor{LightCyan} &  \textbf{78.24}\cellcolor{LightCyan} / 52.56\cellcolor{LightCyan} & 46.01\cellcolor{LightCyan} / 43.22\cellcolor{LightCyan} & \textbf{49.72}\cellcolor{LightCyan} / \textbf{45.73}\cellcolor{LightCyan} & \textbf{57.99}\cellcolor{LightCyan} / \textbf{47.17}\cellcolor{LightCyan}  \\

            \toprule[1pt]
            \bottomrule[1pt]

            \multirow{4}{*}{nuScenes $\rightarrow$ KITTI} 
            & SN & 54.29 / 25.81 & 21.43 / 15.09 & 5.13 / 2.31 & 26.95 / 14.40 \\

            & ST3D & 68.21 / 43.44 & 21.30 / 15.96 &  7.33 / 5.08 & 32.28 / 21.49 \\
            
            & ST3D++ & 68.29 / 40.85 & 22.16 / 16.31 &  7.32 / 3.91 & 32.59 / 20.36 \\

            
            & \textbf{\textsc{ReDB}} \cellcolor{LightCyan} & \textbf{69.76}\cellcolor{LightCyan} / \textbf{46.17}\cellcolor{LightCyan} & \textbf{25.01}\cellcolor{LightCyan} / \textbf{18.88}\cellcolor{LightCyan} & \textbf{9.54}\cellcolor{LightCyan} / \textbf{6.15}\cellcolor{LightCyan} & \textbf{34.77}\cellcolor{LightCyan} / \textbf{23.73}\cellcolor{LightCyan}  \\
            \toprule[0.8pt]
        \end{tabular}
    }
    \end{small}
    \label{tab:hard_compare}
    \vspace{-2ex}
\end{table*}

\section{Experiments}
\subsection{Experimental Setup}
\noindent\textbf{Datasets.} 
Experiments are carried out on three widely used LiDAR 3D object detection datasets: KITTI \cite{DBLP:conf/cvpr/GeigerLU12}, Waymo \cite{DBLP:conf/cvpr/SunKDCPTGZCCVHN20}, and nuScenes \cite{DBLP:conf/cvpr/CaesarBLVLXKPBB20}. We follow \cite{DBLP:conf/cvpr/YangS00Q21, yang2022st3d++} to address different realistic scenarios of 3D domain adaptation: (i) Adaptation across domains with object shifts (\textit{e.g.}, scale, point density), (ii) Adaptation across domains with environmental shift (\textit{e.g.}, data generation regions, LiDAR beams, angels and range).

\noindent\textbf{Baseline Methods.}  
We compare the proposed method with baselines using voxel-based backbone (\textit{e.g.}, SECOND). (i) \textsc{Source Only} refers to the pre-trained model from the source domain evaluated directly on the target domain without adaptations; (ii) SN \cite{DBLP:conf/cvpr/WangCYLHCWC20} takes size statistics of target objects and rescale source objects during pre-training; (iii) ST3D \cite{DBLP:conf/cvpr/YangS00Q21} is the pioneering pseudo labelling-based method that does not require any knowledge about backbone models and dataset statistics; (iv) ST3D++ \cite{yang2022st3d++} is the extended version of ST3D by domain-speciﬁc batch normalization \cite{DBLP:conf/cvpr/ChangYSKH19}, achieving the SOTA performance. (v) \textsc{Oracle} means a fully supervised model trained on the target domain. We also compare with baselines that only work for the point-based backbone (\textit{e.g.}, PointRCNN). (vi) $\textsc{SF-UDA}^\textsc{3D}$ \cite{DBLP:conf/3dim/SaltoriLS0G20} seeks the  best scale for pseudo-labeling from consecutive frames by estimating motion coherence. (vii) \textsc{MLC-Net} \cite{DBLP:conf/iccv/LuoCZZZYL0Z021} is a mean-teacher \cite{DBLP:conf/nips/TarvainenV17} framework aiming to stable optimize the self-training. 

\noindent\textbf{Implementation Details.} Unlike the previous approach, we pre-trained and self-trained a single 3D detection backbone for all categories, simultaneously, rather than unfairly train a single-class model. Our code is built on the point cloud detection codebase OpenPCDet \cite{openpcdet2020} and runs on two NVIDIA Tesla V100 GPUs. For both backbones, we set the batch size to 4 for each GPU. We set hyperparameter $\delta_{\operatorname{cde}}=0.6$, $\delta_{\operatorname{pos}}=0.6$, $\delta_{\operatorname{obc}}=0.3$, $d=5$ and epochs to 90 for all tasks. Regarding the class-balanced sampling, we set $S^r=5$ and $S^g=10$, then increase $S^r$ and decrease $S^g$ by $2$, at each round of pseudo label generation. The implementation code will be available. To fairly evaluate baselines and the proposed method, we utilize the KITTI evaluation metric for evaluation on the three classes: car (equivalent to the vehicle for similar classes in the Waymo), pedestrians and cyclists (equivalent to bicyclist and motorcyclist in nuScenes). Based on the official KITTI evaluation metric, we report the average precision for each class, in both 3D (\textit{i.e.}, $\text{AP}_{\text{3D}}$) and the bird’s eye view (\textit{i.e.}, $\text{AP}_{\text{BEV}}$) over 40 recall positions, with IoU threshold 0.7 for cars and 0.5 for pedestrians and cyclists. We also calculate the mean AP for all classes in 3D and BEV, denoted as $\text{mAP}_{\text{3D}}$ and $\text{mAP}_{\text{BEV}}$, respectively.

\begin{table*}[h] 
\centering 
\caption{Performance comparisons ($\text{AP}_{\text{BEV}}$ and $\text{AP}_{\text{3D}}$) with PointRCNN backbone on nuScenes $\rightarrow$ KITTI task. \textcolor{red}{${\ddagger}$} indicates the results reported in the original paper, where only a single category was adapted.}
\resizebox{1\linewidth}{!}{%
\begin{tabular}{l l | ccc ccc c c c | c c c c c}
\toprule 
& &\multicolumn{3}{c}{\textsc{Car}}&\multicolumn{3}{c}{\textsc{Pedestrian}}&\multicolumn{3}{c}{\textsc{Cyclist}} & \multicolumn{3}{c}{\textsc{Average}} \\ 
\cmidrule(l){3-5}\cmidrule(l){6-8} \cmidrule(l){9-11} \cmidrule(l){12-14}  
 & Method &\textsc{Easy} &\textsc{Mod.} &\textsc{Hard} &\textsc{Easy} &\textsc{Mod.} &\textsc{Hard} &\textsc{Easy} &\textsc{Mod.} &\textsc{Hard} &\textsc{Easy} &\textsc{Mod.} &\textsc{Hard}\\ 
\midrule
\parbox[t]{2mm}{\multirow{7}{*}{\rotatebox[origin=c]{90}{3D}}} 

& \textsc{Source Only} & 42.77 & 32.11 & 28.75 & 43.26 & 37.29 & 33.16 & 3.28 & 4.09 & 4.18 & 29.77 & 24.60 & 22.03 \\
& \textsc{SN} & 66.56 & 50.32 & 45.92 & 42.96 & 37.15 & 32.45 & 9.07 & 7.57 & 7.42 & 39.53 & 31.68 & 28.60\\
& \textsc{ST3D} & 48.85 & 41.90 & 38.92 & 45.67 & 38.71 & 33.09 & 26.50 & 19.35 & 18.38 & 40.34 & 33.32 & 30.13 \\
& \textsc{ST3D++} & 60.45 & 49.36 & 45.88 & 50.77 & 42.43 & 36.64 & 27.20 & 17.94 & 16.52 & 46.14 & 36.58 & 33.01 \\ 
& $\textsc{SF-UDA}^\textsc{3D}$ \textcolor{red}{${\ddagger}$}  & 68.80 & 49.80 & 45.00 & - & - & - & - & - & -  & - & - & - \\
& \textsc{MLC-Net} \textcolor{red}{${\ddagger}$} & 71.30 & 55.40 & 49.00 & - & - & - & - & - & - & - & - & -  \\
\rowcolor{LightCyan}
& \textbf{\textsc{ReDB}} & \textbf{71.45} & \textbf{57.9} & \textbf{53.91} & \textbf{52.32} & \textbf{44.33} & \textbf{37.95} & \textbf{45.13}           & \textbf{32.93}          & \textbf{31.05}           & \textbf{56.30}           & \textbf{45.05}          & \textbf{40.97}           \\

\midrule
\parbox[t]{2mm}{\multirow{4}{*}{\rotatebox[origin=c]{90}{BEV}}} 

& \textsc{Source Only} & 80.91 & 60.75 & 56.00 & 45.84 & 39.78 & 35.45 & 3.38 & 4.45 & 4.57 & 43.37 & 34.99 & 32.00 \\
& \textsc{SN} & 81.17 & 63.34 & 56.80 & 44.80 & 38.73 & 34.39 & 9.65 & 9.39 & 8.93 & 45.21 & 37.15 & 33.67 \\
& \textsc{ST3D} & 85.61 & 69.48 & 64.52 & 47.31 & 39.96 & 35.05 & 32.60 & 23.06 & 21.73 & 55.17 & 44.17 & 40.43 \\
& \textsc{ST3D++} & 85.81 & 69.17 & 64.74 & 52.68 & 45.10 & 39.30 & 31.30 & 20.29 & 18.42 & 56.59 & 44.85 & 40.82 \\
\rowcolor{LightCyan}
& \textbf{\textsc{ReDB}}                & \textbf{91.50}         & \textbf{76.01}          & \textbf{71.42}                    & \textbf{54.83}           & \textbf{45.87}          & \textbf{39.34}           & \textbf{48.44}           & \textbf{35.49}          & \textbf{33.23}           & \textbf{64.93}           & \textbf{52.46}          & \textbf{48.00}           \\
\bottomrule 
\end{tabular}
}
\label{tab:generic_applied} 
\end{table*}

\begin{table*}[t]
  \caption{Ablative study of different modules on adapting Waymo to KITTI task.\vspace{-1ex}}
  \label{tab:abla}
  \centering
  \begin{tabular}{cccc ccc ccc}
    \toprule
     & & & & \multicolumn{3}{c}{\textsc{3D Detection} mAP} &\multicolumn{3}{c}{\textsc{BEV Detection} mAP}\\
     \cmidrule(l){5-7}\cmidrule(l){8-10}
     \textsc{GT} & \textsc{PS} & \textsc{CDE} & \textsc{OBC} &\textsc{Easy} &\textsc{Moderate} &\textsc{Hard} &\textsc{Easy} &\textsc{Moderate} &\textsc{Hard}\\
    \midrule
     -                           &-         &- &-
    &$48.41$ &$42.26$ &$40.18$ &$60.68$ &$53.52$ &$51.01$\\
        $\surd$                       &-         &- &-
         
    &$54.10$ &$45.21$ &$42.42$ &$63.13$ &$56.73$ &$54.00$\\
         -                           &$\surd$        &- &-
    &$51.92$ &$46.04$ &$43.68$ &$62.57$ &$54.57$ &$51.32$ \\
     -                           &$\surd$        &$\surd$ &-
     &$54.04$ &$47.07$ &$44.90$ &$64.70$ &$58.15$ &$55.69$\\
    -  &$\surd$        &- &$\surd$ 
    &$53.99$ &$47.66$ &$44.65$ &$65.72$ &$58.51$ &$55.36$\\
    \rowcolor{LightCyan}
    $\surd$                       &$\surd$   &$\surd$  &$\surd$ &$\textbf{56.52}$ &$\textbf{50.10}$ &$\textbf{47.17}$ &$\textbf{67.36}$ &$\textbf{61.14}$ &$\textbf{57.99}$\\
  \bottomrule
  \bottomrule
\end{tabular}
\vspace{-2ex}
\end{table*}

\subsection{Main Results and Analysis}\vspace{-0.5ex}
We conducted comprehensive experiments on three different 3D adaptation tasks under a multi-class training setting, as summarized and reported in Tab \ref{tab:SOTAcomparison}. We can clearly observe that \textsc{ReDB} consistently improves the performance on Waymo $\rightarrow$ KITTI and nuScenes $\rightarrow$ KITTI by a large margin of 21.27\% and 17.13\% in terms of $\text{mAP}_{\text{3D}}$, which largely close the performance gap between \textsc{Source Only} and \textsc{Oracle}. When comparing with the latest SOTA method ST3D++, our \textsc{ReDB} exhibits superior performance in terms of $\text{mAP}_{\text{BEV}}$ for all three adaptation tasks, with improvements of 2.9\%, 12.5\%, and 12.6\%, respectively. It is worth noting that the proposed \textsc{ReDB} gains significantly more performance for the last two tasks (\textit{i.e.}, Waymo $\rightarrow$ nuScenes and nuScenes $\rightarrow$ KITTI) than the first task (\textit{i.e.}, Waymo $\rightarrow$ KITTI), indicating the \textsc{ReDB} is more effective for adapting to 3D scenes with larger environmental gaps. It is further evident that the \textsc{ReDB} method stands out for its well-balanced performance across all categories, while all baselines are biased towards the most frequently appearing class (\textit{i.e.} car) and underperformed in rare classes (\textit{i.e.} pedestrian and cyclist). Overall, the proposed \textsc{ReDB} excels all baselines on both $\text{mAP}_{\text{3D}}$ and $\text{mAP}_{\text{BEV}}$ across all scenarios of 3D adaptation tasks.

\noindent \textbf{Performance Analysis under Difficult Conditions.} To corroborate the effectiveness of the \textsc{ReDB} approach against the baseline methods, we present additional experimental results in Tab \ref{tab:hard_compare}, using a more challenging evaluation metric: the average precision (AP) at \textbf{Hard} difficulty level defined by KITTI dataset. It is observed that the proposed \textsc{ReDB} outperforms the SOTA baseline ST3D++ by 5.81\% of $\text{mAP}_{\text{3D}}$ for adapting from Waymo to KITTI. Concerning a more challenging adaptation task (\textit{i.e.}, nuScenes → KITTI) involving significant environmental shifts in beam numbers, angles and the point cloud range, the proposed ReDB exhibits superior performance over all other baseline models across all categories: 16.55\% of $\text{mAP}_{\text{3D}}$ than the SOTA baseline. Thus, our approach surpasses the baselines by a large margin when assessed with KITTI metric at \textbf{Hard} difficulty, indicating the efficacy of the \textsc{ReDB} in facilitating effective generalization of 3D detectors to difficult target objects.

\subsection{Component Analysis} \vspace{-1ex}
\begin{figure*}[t]
    \centering
    \includegraphics[width=1\linewidth]{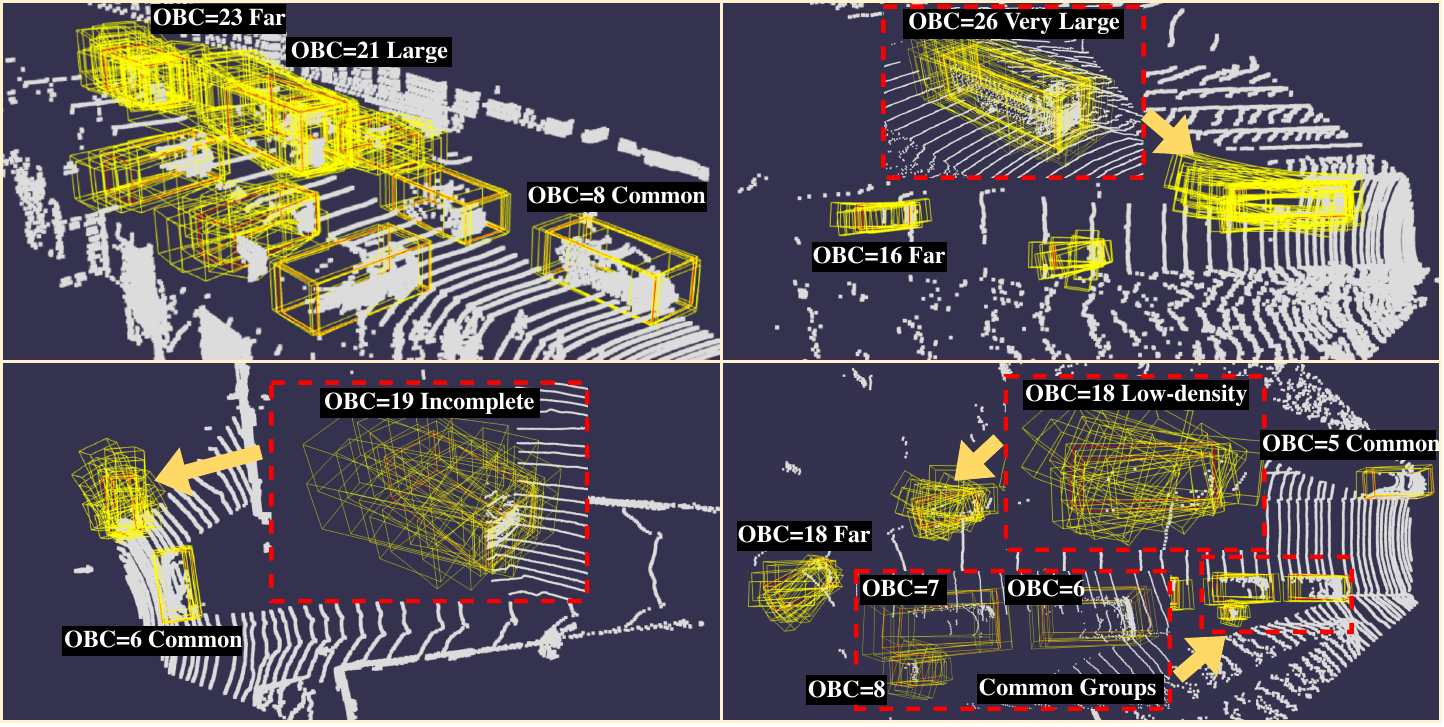}
    \caption{Case studies of the overlapped boxes counting. We visualize the predicted boxes before NMS (in yellow) and its count, with the corresponding geometric description. By inverse frequency sampling of OBCs, geometrical diverse objects are added to the \textsc{ReD}  Pool.}
    \label{fig:vis}
\end{figure*}
\noindent\textbf{Ablation Study.} To investigate the impact of the derived CDE-based reliable pseudo label generation, OBC-based downsampling and class-balanced self-training, we compare five variants of the \textsc{ReDB} model in the task of adapting Waymo to KITTI, shown in Tab \ref{tab:abla}. GT indicates the source ground truth sampling and PS indicates the target pseudo label sampling. For fairness, we keep the same hyperparameters for all ablation variants. To explore the effectiveness of the proposed CDE and OBC modules, we remove each of them and observe a dramatic $\text{mAP}_{\text{3D}}$ drop of 4.9\% (row-5) and 5.34\% (row-4) compared to the full model (row-6), respectively. To validate the proposed class-balanced self-training, we find when comparing to the non-sampling baseline (row-1), the source ground truth sampling (row-2) and the target pseudo label sampling (row-3) increase $\text{mAP}_{\text{3D}}$ of 7\% and 8.9\%, respectively. It is noticeable that, source ground truth sampling (row-2) produces remarkable $\text{mAP}_{\text{3D}}$ gains at easy level (5.69\%) but very limited gains at moderate and hard levels (2.95\% and 2.24\%), caused by a lack of knowledge about diverse geometrical features in the target domain. Therefore, removing any module from the proposed \textsc{ReDB} induces a clear drop in 3D / BEV mAP scores, confirming the importance of each component that contributes to multi-class 3D domain adaptation. 

\noindent\textbf{Sensitivity to Detector Architecture.}  To validate the sensitivity of performance to choices of voxel-based and point-based detectors, we plug the proposed \textsc{ReDB} paradigm into PointRCNN. As reported in Tab \ref{tab:generic_applied}, our approach consistently outperforms the latest SOTA method ST3D++ by 23.32\% and 24.11\% regarding $\text{mAP}_{\text{3D}}$ at moderate and hard levels, respectively. Notably, our multi-class method has even achieved superior performance (10.02\% and 19.8\%) compared to \textsc{MLC-Net} and $\textsc{SF-UDA}^\textsc{3D}$, both of which were under the unfair single-class training setting and are exclusively designed for the point-based 3D detector. The outstanding results of the proposed method using a variety of 3D detectors, demonstrate the \textsc{ReDB} is a model-agnostic and insensitive to detector architecture.

\begin{figure}[t]
    \centering
    \includegraphics[width=1\linewidth]{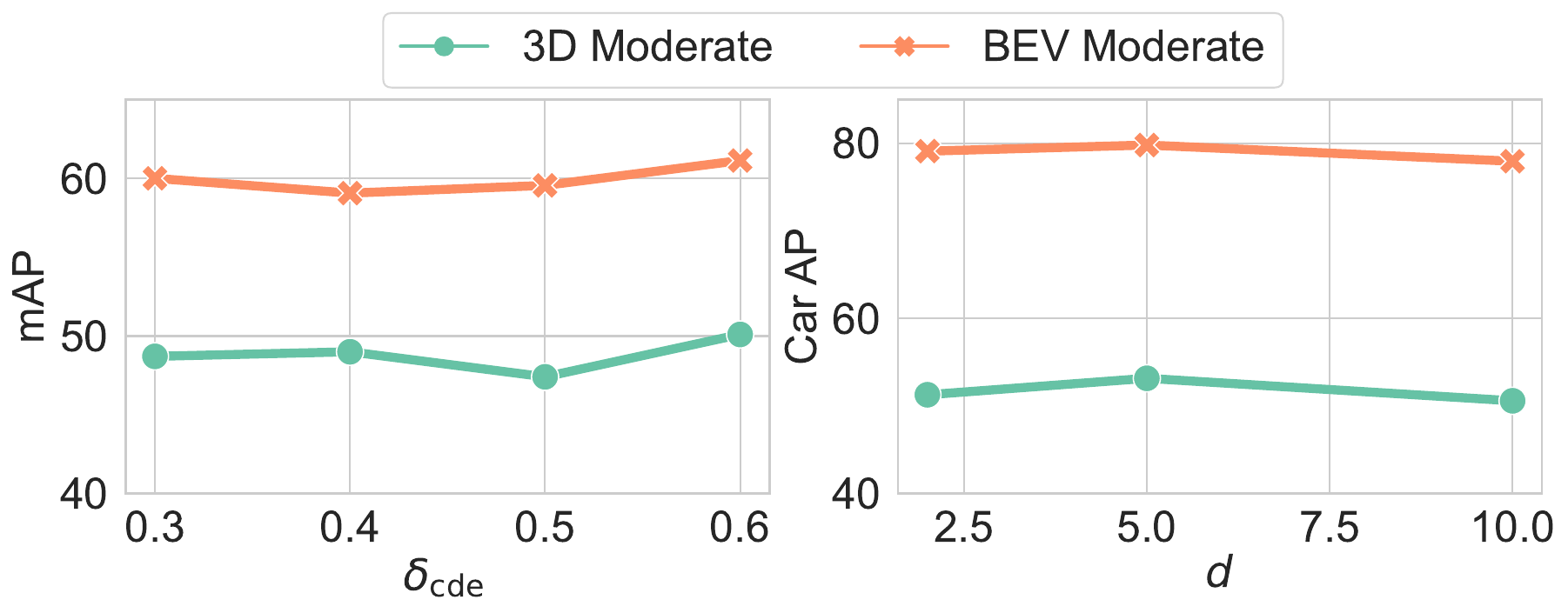}
    \caption{Hyperparameter sensitivity of $\delta_{\operatorname{cde}}$ and $d$.}
    \label{fig:parameter_sensitivity}
\end{figure}

\noindent\textbf{Sensitivity to the hyperparameters $\delta_{\operatorname{cde}}$ and $d$.} We show the sensitivity of our approach to varying the CDE IoU threshold $\delta_{\operatorname{cde}}$ and diversity downsampling rate $d$ in Fig \ref{fig:parameter_sensitivity}. We vary the value of $\delta_{\operatorname{cde}}$ and $d$, from $[0.3, 0.6]$ and $[2, 10]$, respectively, while fixing the other configurations to the default setting. We can observe that, when $\delta_{\operatorname{cde}}$ varies, there is only 2.69\% and 2.07\% fluctuation on $\text{mAP}_{\text{3D}}$ and $\text{mAP}_{\text{BEV}}$ scores, respectively. Regarding the variation of $d$, we notice when $d$ is set as a large value (\textit{e.g.}, $d=10$), the performance drops within an acceptable range: 1.37\% in $\text{mAP}_{\text{3D}}$ and 2.57\% in $\text{mAP}_{\text{BEV}}$. Such a small drop could be caused by the insufficient scale of the down-sampled \textsc{ReD} pool. Hence, selecting a proper value of downsampling rate (\textit{e.g.}, $d=5$), reaches the best results. Little fluctuation caused by varying $\delta_{\operatorname{cde}}$ and $d$ evidence that, the proposed method is robust to different choices of hyperparameters.



  



\vspace{1ex}
\noindent\textbf{Visualization of OBC-based Diversity.}\label{sec:vis_OBC} To intuitively understand the merits of our proposed diversity module, Fig \ref{fig:vis} visualizes the counted predicted boxes before NMS for different pseudo-labeled target objects. We can see that most objects with small OBC values (\textit{e.g.}, between 5 and 8) are, (1) typically closer to the LiDAR sensor, (2) with a complete object shape, (3) and commonly small-sized objects. These objects usually have highly similar and complete geometric features, making up the majority of the dataset. In contrast, objects with high OBCs are usually diverse in one or more aspects of geometrical representations. Regarding the object size, large-sized objects tend to yield high OBC scores (\textit{e.g.}, 21 and 26). In addition to object volume, we can also find objects that are obviously far away from the LiDAR center can produce higher OBC values (from 16 to 23), and low-density and heavily occluded objects also have high OBC values of 18 and 19, respectively. More statistical analysis can be found in supplementary materials. Therefore, the proposed OBC metric can effectively quantify the diversity of pseudo-labels in multiple dimensions of geometric features, aiding the 3D detector to learn a more diverse distribution of target objects, thereby alleviating multidimensional object shift.



\section{Conclusion}

This paper revisits the setting of unsupervised domain-adaptive 3D detection and investigates the generation of reliable, diverse, and class-balanced pseudo labels for effective multi-class adaptation. The quantitative and qualitative analysis demonstrates the effectiveness of the derived cross-domain examination and OBC-based downsampling strategies, allowing pseudo labels that are robust to environmental and object shifts to be preserved. The class-balanced self-training helps detectors to learn frequently appearing and rare classes simultaneously. The proposed approach is proven to be versatile and can accommodate voxel-based and point-based 3D detectors.

\section{Acknowledgments}
This work was supported by Australian Research Council (DP230101196, CE200100025).

{\small
\bibliographystyle{ieee_fullname}
\bibliography{mainbib}
}

\end{document}


\title{Supplementary Material for \\ Revisiting Domain-Adaptive 3D Object Detection by Reliable, Diverse and Class-balanced Pseudo-Labeling}

\author{Zhuoxiao Chen$^{1}$ \; Yadan Luo$^{1}$  \; Zheng Wang$^{2}$ \; Mahsa Baktashmotlagh$^{1}$ \;  Zi Huang$^{1}$ \\
$^1$The University of Queensland\quad $^2$University of Electronic Science and Technology of China \\
{\tt\small \{zhuoxiao.chen, y.luo, m.baktashmotlagh, helen.huang\}@uq.edu.au, zh\_wang@hotmail.com } \\
}

\maketitle
\ificcvfinal\thispagestyle{empty}\fi
In this supplementary material, we provide a detailed description of the proposed algorithm (Sec \ref{sec:alg}), a conceptual comparison with previous method (Sec \ref{sec:cc}) and additional experimental results and analysis (Sec \ref{sec:add_exp}). 

\renewcommand{\algorithmicrequire}{\textbf{Input:}}
\renewcommand{\algorithmicensure}{\textbf{Output:}}
\begin{algorithm*}
\caption{The algorithm of ReDB for domain-adaptive 3D object detection}\label{alg:ReDB}
\begin{algorithmic}
\Require{\\
$\{(X^s_i, Y^s_i)\}^{N_s}_{i=1}$ - source point clouds with human annotations, where all labelled boxes: $\{b_j\}^{B}_{j=1}=\{b_j | b_j\in Y_i^s\}_{i=1}^{N_s}$   \\
$\{X^t_i\}^{N_t}_{i=1}$ - target point clouds without human annotations \\
$F(\cdot)$ - the 3D object detector \\
$f_{\operatorname{ROS}}(\cdot)$ - the random object scaling funcion\\
$E$ - total number of training epochs \\
$L$ - a list of epochs requiring pseudo labelling
}
\Ensure{\\$F(\cdot)$ the 3D detector adapted to the target domain}
\\\hrulefill
\State \textcolor{gray}{/* Stage 1: Pretrain on the Source Domain */}
\State $F(\cdot) \gets \{(X^s_i,f_{\operatorname{ROS}}( Y^s_i))\}^{N_s}_{i=1}$
\For{$e \in [1,\cdots , E]$}
\If{$e=1 \,\,\text{or}\,\, e \in L$} 
\State \textcolor{gray}{/* Stage 2: Generate Pseudo Labels on the Target Domain */}
\State $\{\widehat{Y}_i^t\}^{N_t}_{i=1} \gets F(\{X^t_i\}^{N_t}_{i=1})$
\If{$e=1$} \Comment{Cross-domain Consistency Examination via Equation (2), (3)}
\State $\{\widehat{Y}_i^t\}^{N_t}_{i=1} \gets \operatorname{Reliability}(\{(X^t_i, \widehat{Y}_i^t)\}^{N_t}_{i=1}, \{(X^s_i, Y^s_i)\}^{N_s}_{i=1})$
\EndIf
\State $\{\hat{b}_j\}^{\hat{B}}_{j=1} = \{\hat{b}_j | \hat{b}_j\in\hat{Y}_i^t\}_{i=1}^{N_t}$
\State $\{\hat{b}_j\}^{\hat{B}/d}_{j=1} \gets \operatorname{Diversity}(\{\hat{b}_j\}^{\hat{B}}_{j=1})$ \Comment{OBC-based Downsampling via Equation (4), (5)}
\State $ \{(X^t_i, \widehat{Y}_i^t)\}^{N_t}_{i=1} \gets \operatorname{Balance}(\{(X^t_i, \widehat{Y}_i^t)\}^{N_t}_{i=1}, \{\hat{b}_j\}^{\hat{B}/d}_{j=1},f_{\operatorname{ROS}}(\{b_j\}^{B}_{j=1}))$ 
\State \Comment{Class-balanced Sampling via Equation (6), (7), (8)}
\EndIf
\State \textcolor{gray}{/* Stage 3: Self-train on the Target Domain */}
\State $F(\cdot) \gets \{(X^t_i, \widehat{Y}_i^t)\}^{N_t}_{i=1}$
\EndFor
\end{algorithmic}
\end{algorithm*}

\begin{table*}[t]
  \caption{Comparison of prior Domain Adaptive 3D Detection methods in two aspects: applicability and addressed domain shifts. \textbf{The main difference is that the existing methods has constrained applicability, and address incomprehensive domain shifts.}}
  \label{tab:compare}
  \centering
  \resizebox{1\linewidth}{!}{%
  \centering
  \begin{tabular}{c | cccc | ccc}
    \toprule
     & \multicolumn{4}{c}{Applicability} &\multicolumn{2}{c}{Addressed Domain Shift}\\
     \cmidrule(l){2-5}\cmidrule(l){6-7}
     \textsc{Method} & Multi-class & Compatible & UDA & Space Complexity & Object Shift & Environmental Shift \\
    \midrule
    SN \cite{DBLP:conf/cvpr/WangCYLHCWC20}  &\ding{52} &\ding{52} &\textcolor{red}{\ding{56}} &\ding{52} &Scale &\textcolor{red}{\ding{56}}  \\
    \textsc{LiDAR Distil} \cite{DBLP:conf/eccv/WeiWRLZL22} &\textcolor{red}{\ding{56}} &Grid-based &\textcolor{red}{\ding{56}} & Multiple models &\textcolor{red}{\ding{56}} &Beam number  \\
    \textsc{MLC-Net} \cite{DBLP:conf/iccv/LuoCZZZYL0Z021}
    &\textcolor{red}{\ding{56}} &Point-based &\ding{52} &Dual models &Scale &\textcolor{red}{\ding{56}} \\
    ST3D \cite{yang2022st3d++}          
    &\textcolor{red}{\ding{56}} &\ding{52} &\ding{52} &Memory bank &Scale &\textcolor{red}{\ding{56}}  \\
    ST3D++ \cite{DBLP:conf/cvpr/YangS00Q21}      
    &\textcolor{red}{\ding{56}} &\ding{52} &\ding{52} &Memory bank &Scale &\textcolor{red}{\ding{56}} \\
    \midrule
    \textsc{ReDB} &\ding{52} &\ding{52} &\ding{52} &\ding{52} &
    \begin{tabular}{@{}c@{}} Diversity geometrics \\ (scale, density, distance, etc) \end{tabular}
    &    \begin{tabular}{@{}c@{}} All aspects \\ (beam number, angle, etc) \end{tabular}
      \\
  \bottomrule
\end{tabular}
}
\end{table*}



\section{Algorithm}\label{sec:alg}
To thoroughly describe the procedure of unsupervised domain-adaptive 3D object detection by the proposed \textsc{ReDB}, we present the Algorithm \ref{alg:ReDB}.  In the stage one, we pretrain the 3D detector $F(\cdot)$ on the source domain with the randomly scaled objects \cite{DBLP:conf/cvpr/YangS00Q21} $\{(X^s_i,f_{\operatorname{ROS}}( Y^s_i))\}^{N_s}_{i=1}$. If the current epoch is initial $e=1$ or in the list $L=[31, 61, 91]$, which specifies the epochs requiring pseudo-labelling, we enter stage 2 to generate reliable, diverse and balanced \textsc{ReDB} pseudo labels. We first inference the target domain $\{X^t_i\}^{N_t}_{i=1}$ and obtain the pseudo label $\{\widehat{Y}_i^t\}^{N_t}_{i=1}$ for each target point cloud. Then, if the current epoch $e=1$, which means the 3D detector has no knowledge about the target domain, the pseudo label might be unreliable thus we need examination. We feed the pseudo label paired with the corresponding point clouds $\{(X^t_i, \widehat{Y}_i^t)\}^{N_t}_{i=1}$ and the source pairs $\{(X^s_i, Y^s_i)\}^{N_s}_{i=1}$ into the proposed Cross-domain Consistency Examination (CDE) module that filters out the unreliable pseudo labels via Equation (2), (3). Once the reliability of pseudo labels is guaranteed, we target at obtaining a more geometrically diverse subset of pseudo-labeled boxes $\{\hat{b}_j\}^{\hat{B}/d}_{i=j}$ to close the object gap from different geometrical aspects, using OBC-based downsampling via Equation (4), (5). Next, to alleviate the class imbalance, we sample class-balanced reliable and diverse pseudo labels $\{\hat{b}_j\}^{\hat{B}/d}_{j=1}$ and randomly scaled source labels $f_{\operatorname{ROS}}(\{b_j\}^{B}_{j=1})$ to each point cloud $\{X^t_i, \widehat{Y}_i^t\}^{N_t}_{i=1}$ via Equation (6), (7). At the end of stage 2, we have a subset of \textsc{ReDB} pseudo labels. We train this subset in stage 3 for several epochs until the current epoch appears in the pseudo-labelling epoch list $L$. The algorithm is then alternating between stage 2 and stage 3 until running out of all epochs.

\section{Conceptual Comparison}\label{sec:cc}
This section establishes a connection between the proposed \textsc{ReDB} and previous domain-adaptive 3D detection approaches, with emphasis on real-world applicability and aspects to address domain shifts, as summarized in Tab \ref{tab:compare}. 

\noindent\textbf{Statistical Normalization (SN) \cite{DBLP:conf/cvpr/WangCYLHCWC20}:} A data modification approach that modifies the object size of the source domain to match the target statistics, in order to address scale-induced object shift. Nonetheless, this method demands the access to object statistics of the target domain, rendering it unfeasible in practical scenarios where domain knowledge is not available.

\noindent\textbf{\textsc{MLC-Net} \cite{DBLP:conf/iccv/LuoCZZZYL0Z021}:} 
A teacher-student paradigm, in which the teacher parameters are aggregated by an exponential moving average (EMA) of the student model and updated iteratively. The teacher is in charge of producing pseudo labels that supervise the learning of the student model.  The technique leverages the historical weights to predict smooth pseudo labels for self-training on the target domain. However, this method suffers from four drawbacks: (1) it overlooks the environmental gap, leading to erroneous pseudo-labeling during the initial pseudo-label generation stage. However, our proposed cross-domain examination (CDE) offers a simple solution to this problem. (2) The employment of two models (student and teacher) simultaneously doubles the training time and GPU memory consumption. Additionally, the point-wise consistency loss that supervised the learning of student model yields more gradient back-propagation, which further increases the training time. (3) Despite being compatible with point-based 3D detectors, this method cannot be applied to mainstream grid-based detectors. (4) This approach only tailors the model training and inference for a single class, which is not a viable strategy in real-world situations.

\noindent\textbf{\textsc{LiDAR Distil} \cite{DBLP:conf/eccv/WeiWRLZL22}:} 
A teacher-student framework that seeks to enhance the generalizability of 3D detectors to the domain with different beam numbers. In particular, the teacher and student networks are trained using high-beam and low-beam data, respectively. The student network is subsequently trained to align the regions of interest on the bird eye vew (BEV) feature maps with those of the teacher model. The primary objective of this approach is to mitigate the environmental gap that stems from inconsistent beam numbers. However, this method encounters similar shortcomings as \textsc{MLC-Net} or SN, including the prerequisite knowledge about the target domain (\textit{i.e.}, beam number), the employment of multiple teacher networks leading to considerable computation time and GPU memory consumption, restricted applicability to point-based 3D detectors, and unfair single-class adaptation. Additionally, this study fails to account for shifts in objects and other environmental factors, such as disparities in beam angle, range, and data collection locations and time.

\begin{table}[ht!]
    \definecolor{Gray}{gray}{0.9}
    \newcolumntype{a}{>{\columncolor{Gray}}c}
    \centering
    \begin{small}
        \begin{tabular}{lcca}
            \bottomrule[1pt]
             $S^r$ & $S^g$ & {$S^\Delta$} & mAP$_{\text{BEV}}$ / mAP$_{\text{3D}}$  \\
            \hline
            0 & 10  & 0 & 56.73 / 45.21  \\
            0 & 10  & 2 & 60.27 / 49.25  \\
            10 & 20 & 2 & 60.22 / 48.73  \\
            5 & 10 & 2  & \textbf{61.14} / \textbf{50.10} \\
            \toprule[0.8pt]
        \end{tabular}
    \end{small}
    \caption{Sensitivity analysis for $S^r$, $S^g$ and $S^\Delta$.}
    \label{tab:supp_sensi_anal}
\end{table}

\noindent\textbf{ST3D \cite{DBLP:conf/cvpr/YangS00Q21} / ST3D++ \cite{yang2022st3d++}:} 
A self-training strategy that employs random object scaling (ROS) to pre-train the source data and a memory bank to store and update all pseudo labels for self-training the target data. The objective of ROS is to alleviate object shift by allowing the pre-trained detector to recognize objects with a wide range of scales. The memory bank combines historical and current pseudo labels to generate more consistent pseudo labels. ST3D++ is an extended version that incorporates a domain-specific batch normalization \cite{DBLP:conf/cvpr/ChangYSKH19} with the aim of disentangling the statistic estimation (\textit{i.e.}, mean and variance) in different domains within batch normalization layers. The benefit of this series of work (\textit{i.e.}, ST3D and ST3D++) over prior works is that, they do not necessitate any prior knowledge of the target domain and are not constrained by detector types. Despite some progress, significant challenges persist in several aspects, including neglecting environmental shifts, unfair single-class adaptation, and excessive storage requirements for retaining historical pseudo labels.

\noindent\textbf{\textsc{ReDB}:} Tab \ref{tab:compare} highlights the advantages offered by the proposed \textsc{ReDB}, concerning the approach applicability and handled domain shifts. The proposed \textsc{ReDB} (1)  facilitates multi-class adaptation via the proposed class-balanced self-training, (2) is compatible with all types of point cloud encoders used in modern 3D detectors, (3) employs an unsupervised domain adaptation (UDA) approach, which eliminates the need for any prior knowledge about the target domain, (4) does not incur extra costs for GPU memory and disk storage, (5) lastly, provides a solution to comprehensively identify and address both object shifts and environmental shifts.

\begin{figure*}[t]
    \centering
    \includegraphics[width=1\linewidth]{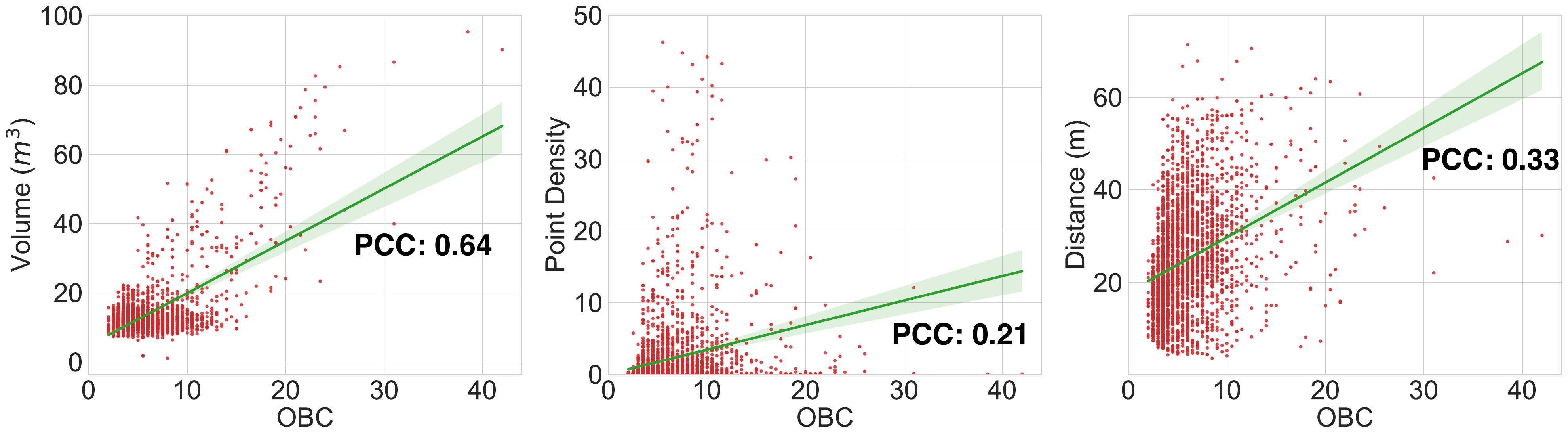}
    \caption{Statistical analysis on overlapped boxes counting (OBC). Each red point indicates a pseudo-labeled box associated with its OBC value and 1-dimensional geometrical feature (\textit{e.g.}, volume, point density and distance). The green line is the regression line, and the translucent band indicates the size of the confidence interval for the regression estimate. We calculate the Pearson correlation coefficient (PCC) to show the correlation between OBC value and different geometrical features.}
    \label{fig:obc_analysis}
\end{figure*}

\section{Additional Experiments}\label{sec:add_exp}
\subsection{Implementation Details}
\noindent\textbf{Hyperparameter settings.} For self-training on the target domain, we set the total training epochs $E$ as $120$ and pseudo label generations list $L$ as $[31, 61, 91]$. Thus, we generate the pseudo labels at the initial epoch of self-training, then update pseudo labels every $30$ training epochs. \textbf{We provide complete configuration files for all experiments in our code repository, attached with the supplementary material.}

\noindent\textbf{Fair Model Selection.} 
The prior approaches evaluate the checkpoints based on the performance of the target domain, which is \textbf{not fair and impractical}, because the target labels not observable under the unsupervised domain adaptation (UDA) setting. Hence, We again revisit model selection in a fair manner without accessing any target labels. For selecting the pre-trained models, we simply opt for the one with the \textbf{lowest source risk}. Regarding the self-trained checkpoints, we observe that the modern 3D detectors optimized with Adam Onecycle \cite{DBLP:journals/corr/abs-1803-09820} are typically reaching the best performance at the cycle's end. Based on this observation, we conduct multiple training cycles (each cycle contains $30$ epochs) with the Adam Onecycle optimizer and generate pseudo-labels at the end of each cycle. The final selected model is the \textbf{last checkpoint of the last round} after several rounds of self-training. By doing so, we not only assure the pseudo-labels are generated by optimal models at each round, but also can simply decide the last checkpoint as the final model without knowing the target labels.

\subsection{Sensitivity Analysis of $S^r$, $S^g$ and $S^\Delta$}
In this section, We examine the sensitivity of our approach to different values of  hyperparameters $S^r$, $S^g$ and the increasing and decreasing number $S^\Delta$ for $S^r$ and $S^g$, respectively. Tab \ref{tab:supp_sensi_anal} presents different combinations of $S^r$, $S^g$ and $S^\Delta$. A comparison of the row-1 and row-2 reveals that injecting pseudo-labeled \textsc{ReDB} objects progressively, can yield a notable performance boost, which are 6.24\% and 8.94\% in mAP$_{\text{BEV}}$ and mAP$_{\text{3D}}$, respectively. When comparing row-2, row-3 and row-4, we find that either sampling a large or small number of objects leads to similar performance, fluctuating at 0.87\% mAP$_{\text{BEV}}$ and 1.37 mAP$_{\text{3D}}$, respectively. As a result, to secure an appropriate time complexity, we commonly select the values in row-4 for $S^r$, $S^g$ and $S^\Delta$.

\begin{table}[t]
    \definecolor{Gray}{gray}{0.9}
    \newcolumntype{a}{>{\columncolor{Gray}}c}
    \centering
    \begin{small}
        \begin{tabular}{lcc}
            \bottomrule[1pt]
             Method & nuScenes $\rightarrow$ KITTI &  Waymo $\rightarrow$ nuScenes  \\
            \hline
            ST3D & 25h 24m 48s & 27h 13m 42s  \\
            ST3D++ & 22h 22m 28s  & \textbf{24h 57m 23s}  \\
            \textsc{ReDB}\cellcolor{LightCyan} & \textbf{20h 21m 1s}\cellcolor{LightCyan} & 29h 53m 16s\cellcolor{LightCyan}   \\
            \toprule[0.8pt]
        \end{tabular}
    \end{small}
    \caption{Self-training Time Comparison.}
    \vspace{-0.5cm}
    \label{tab:time}
\end{table}

\subsection{Statistical Analysis on OBC scores}
In this section, we plot the statistical correlation of the proposed overlapped box counting (OBC) with different geometrical features in Fig \ref{fig:obc_analysis}. The first plot shows that there is a clear relationship between OBC and box scales, resulting in a Pearson correlation coefficient (PCC) of 0.64. The second and third plots suggest that point density and object-to-sensor distance have a minor correlation with the OBC score, with PPC 0.21 and 0.33, respectively. Such minor correlations are discovered because the diversity of an object is not solely determined by a single factor (\textit{e.g.,} object density, distance), but an unmeasurable combination of different geometrical features.  To address multi-factor object shifts in 3D scenes, a universal metric is required. Instead of manually designing a linear combination of different factors, the proposed OBC score focuses on reflecting the overall diversity for each 3D object.

\subsection{Time Analysis on Self-training}
To demonstrate that the effectiveness of the proposed \textsc{ReDB} does not benefit from additional extensive computation, we report the runtime of the entire self-training process, including pseudo label generation, in Tab \ref{tab:time}. Note that for all compared approaches, we use two Tesla V100-PCIE-16GB GPUs for the nuScenes $\rightarrow$ KITTI task and one NVIDIA RTX A6000 GPU for the Waymo $\rightarrow$ nuScenes task. The backbone 3D detector is SECOND for all approaches. The time comparison results show that our method takes a similar amount of time as the two existing baseline methods. Specifically, on the task of nuScenes $\rightarrow$ KITTI, we are approximately 2 to 4 hours faster, while 3 to 5 hours slower for the Waymo $\rightarrow$ nuScenes task. This is due to the fact that each single frame of point cloud in Waymo is much larger than that in nuScenes, thus CDE takes longer to infer the pseudo labels pasted to the Waymo point cloud. However, both existing methods rely on the memory bank technique, which is not only time-consuming but also memory-hungry. 

{\small
\bibliographystyle{ieee_fullname}
\bibliography{mainbib}
}